\documentclass[lettersize,journal]{IEEEtran}
\usepackage{multirow}
\usepackage{amsmath,amsfonts}
\usepackage{algorithmic}
\usepackage{algorithm}
\usepackage{array}
\usepackage[caption=false,font=normalsize,labelfont=sf,textfont=sf]{subfig}
\usepackage{textcomp}
\usepackage{stfloats}
\usepackage{url}
\usepackage{verbatim}
\usepackage{graphicx}
\usepackage{cite}
\usepackage{hyperref}
\hypersetup{
	colorlinks=true,
	linkcolor=blue,
	filecolor=blue,      
	urlcolor=black,
	citecolor=blue,
    }

\usepackage{amsmath}
\usepackage{amssymb}
\usepackage{pifont}
\usepackage{threeparttable}
\begin{document}

\title{You Need a Transition Plane: Bridging Continuous Panoramic 3D Reconstruction with Perspective Gaussian Splatting}

\author{Zhijie~Shen, Chunyu~Lin, \IEEEmembership{Member,~IEEE}, Shujuan Huang, Lang Nie,\\ Kang~Liao, \IEEEmembership{Member,~IEEE}, and Yao~Zhao, \IEEEmembership{Fellow,~IEEE}
        % <-this % stops a space
\thanks{Zhijie Shen, Chunyu Lin, Shujuan Huang, Lang Nie, and Yao Zhao are with the Institute of Information
Science, Beijing Jiaotong University, Beijing 100044, China, and also with
Visual Intelligence +X International Cooperation Joint Laboratory of MOE,
Beijing 100044, China (e-mail: zhjshen@bjtu.edu.cn, cylin@bjtu.edu.cn, nielang@bjtu.edu.cn,
yzhao@bjtu.edu.cn).}
\thanks{Kang Liao is with the College of Computing and Data Science, Nanyang Technological University, Singapore (e-mail: kang.liao@ntu.edu.sg, wslin@ntu.edu.sg).}}

\maketitle
\begin{abstract}
Recently, reconstructing scenes from a single panoramic image using advanced 3D Gaussian Splatting (3DGS) techniques has attracted growing interest. Panoramic images offer a 360°×180° field of view (FoV), capturing the entire scene in a single shot. However, panoramic images introduce severe distortion, making it challenging to render 3D Gaussians into 2D distorted equirectangular space directly. Converting equirectangular images to cubemap projections partially alleviates this problem but introduces new challenges, such as projection distortion and discontinuities across cube-face boundaries. To address these limitations, we present a novel framework, named \textbf{TPGS}, to bridge continuous panoramic 3D scene reconstruction with perspective Gaussian splatting. Firstly, we introduce a \textit{Transition Plane} between adjacent cube faces to enable smoother transitions in splatting directions and mitigate optimization ambiguity in the boundary region. Moreover, an intra-to-inter face optimization strategy is proposed to enhance local details and restore visual consistency across cube-face boundaries. Specifically, we optimize 3D Gaussians within individual cube faces and then fine-tune them in the stitched panoramic space. 
Additionally, we introduce a spherical sampling technique to eliminate visible stitching seams. Extensive experiments on indoor and
outdoor, egocentric, and roaming benchmark datasets demonstrate that our approach outperforms existing state-of-the-art methods. Code and models will be available at \url{https://github.com/zhijieshen-bjtu/TPGS}.
\end{abstract}

\begin{figure}[t]
  \centering
  \includegraphics[width=0.46\textwidth]{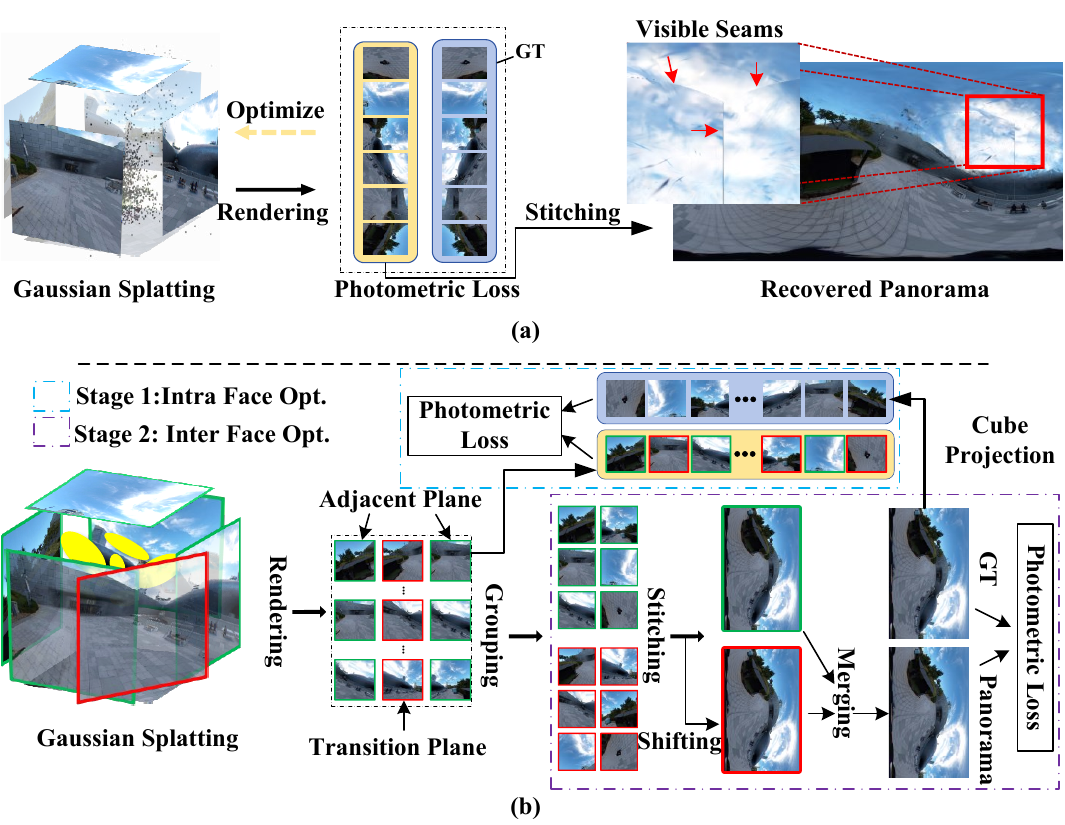} %1.png是图片文件的相对路径 
  \caption{Comparison of the commonly used framework (a)  against 
 our proposed method (b). Our approach introduces a transition plane along with an intra-to-inter face optimization strategy to address the challenges inherent in cubemap projection.} %caption是图片的标题
  \label{fig:moti}
\end{figure}
\section{Introduction}
3D scene reconstruction enables accurate recovery of geometric details and depth, providing a comprehensive and realistic representation of scenes for applications such as virtual reality (VR), augmented reality (AR), robotic navigation, and autonomous driving. However, conventional reconstruction methods based on perspective images have significant limitations. These approaches rely heavily on accurate feature matching\cite{gava2023sphereglue} across images captured from multiple viewpoints. Due to the limited FoV of perspective cameras, precise reconstruction of entire scenes usually requires numerous images, resulting in high computational costs. In contrast, panoramic imaging models effectively capture the entire scene from a single viewpoint\cite{hsu2021moving,huang2022real}, significantly reducing data acquisition and computational overhead \cite{chen2022casual,chiu2023360mvsnet,eder2019pano,liao2023deep}. Moreover, recent advances have enabled 3DGS\cite{kerbl20233d}—a point-based differentiable rendering technique—to achieve efficient optimization and real-time rendering, making it a leading method for 3D scene reconstruction. Combining panoramic imaging with 3DGS integrates their complementary strengths, presenting a promising approach toward accurate and real-time 3D reconstruction.

\begin{figure}[t]
  \centering
  \includegraphics[width=0.46\textwidth]{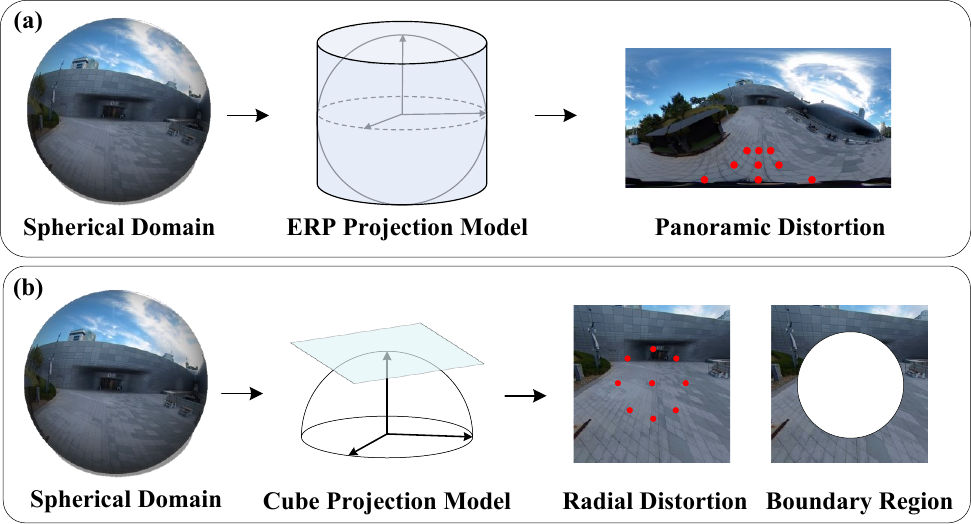} %1.png是图片文件的相对路径 
  \caption{Panoramic distortion \textit{v.s.} projection distortion. (a) The spherical image is projected onto a 2D equirectangular panorama. Non-uniform sampling density in the latitude direction causes panoramic distortion. (b) The spherical image is mapped onto six cube faces. Non-uniform sampling density from the center to the edges causes projection distortion, reducing resolution in boundary regions. This uneven resolution disrupts the uniformity of supervision signals during optimization.
} %caption是图片的标题
  \label{fig:distortion}
\end{figure}

\begin{figure}[t]
  \centering
  \includegraphics[width=0.35\textwidth]{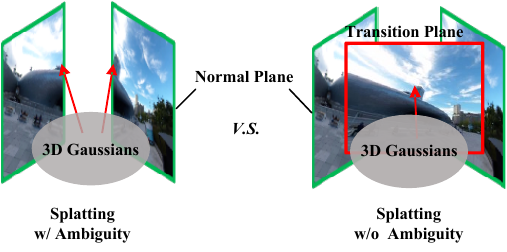} %1.png是图片文件的相对路径 
  \caption{Splatting ambiguity removal via the proposed transition plane. %When 3D Gaussians are splatted onto the boundary region of a cube, they may span across adjacent planes. During optimization, these planes can independently update the same Gaussian’s attributes, leading to optimization inconsistencies and ambiguity. The transition plane method addresses this issue by splatting 3D Gaussians onto an intermediate plane, thereby relieving optimization ambiguity.
  When 3D Gaussians projected onto cube boundaries may span adjacent planes, causing independent updates to the same Gaussian attributes and leading to optimization conflicts and ambiguities. The transition plane method resolves this by splatting Gaussians onto an intermediate plane} %caption是图片的标题
  \label{fig:splam}
\end{figure}

Omnidirectional\footnote{Panoramic and Omnidirectional are equal in this paper.} imaging models represent the whole scene in a distorted way, which exhibits a significant domain gap with the perspective camera models. Consequently, directly applying the original 3DGS method to reconstruct scenes from omnidirectional images typically yields suboptimal results\cite{bai2024360} (as shown in Figure \ref{fig:moti}a). To mitigate the negative impact of panoramic distortion (specifically referring to distortion introduced by the equirectangular representation, Figure \ref{fig:distortion}a), an effective strategy involves converting the equirectangular projection format into cubemap or tangent projections, thereby aligning more naturally with the perspective-based 3DGS framework. This approach partially resolves the domain mismatch that results from differences in camera imaging models. Nevertheless, although cubemap projection alleviates panoramic distortion, it inevitably introduces projection distortion (shown in Figure \ref{fig:distortion}b) within each cube face, resulting in unbalanced supervision (details in Figure \ref{fig:distortion}). Moreover, abrupt changes in viewing directions between adjacent faces also lead to noticeable boundary discontinuities. When 3D Gaussians are splatted onto these regions, ambiguities arise (details in Figure \ref{fig:splam}left), leading to unstable training. On the other hand, some studies\cite{bai2024360,lee2024odgs,li2024omnigs} focus on developing panorama-specific 3DGS frameworks. Specifically, these approaches leverage local approximations based on the tangent plane to bridge the domain gap between panoramic and perspective images. However, this strategy introduces additional approximation errors (beyond the initial errors introduced by approximating 3D Gaussians as 2D Gaussians\cite{matsuki2024gaussian,lee2024odgs}). Furthermore, accurate tangent-plane approximations require sufficiently small local patches, forcing frequent pruning and splitting of 3D Gaussians during optimization, thus significantly increasing GPU memory consumption. Additionally, such panorama-specific strategies fail to directly benefit from recent advancements\cite{huang2024error,huang2024sc} in general-purpose 3DGS techniques. Developing a practical scheme based on the perspective 3DGS framework is necessary.  

In this paper, we present \textbf{TPGS}, a novel framework for 3D scene reconstruction from omnidirectional imagery with the perspective Gaussian splatting technique (shown in Figure \ref{fig:moti}b). To address the challenges introduced by the cubemap projection, we propose a \textit{\textbf{Transition Plane}} approach. Specifically, we insert an intermediate plane between adjacent cube faces to alleviate the uneven image quality near face boundaries and eliminate splatting ambiguity. %achieve smoother transitions in viewing directions and alleviate uneven pixel distributions near face boundaries. 
Consequently, 3D Gaussians in boundary regions can be optimized more effectively by splatting onto this intermediate plane (as shown in Figure \ref{fig:splam}right). However, achieving globally consistent optimization requires stitching the cube faces into an equirectangular panorama. However, this process reintroduces panoramic distortions, which undermines the effectiveness of the transition plane approach.
%However, stitching cube faces into an equirectangular panoramic image reintroduces panoramic distortion, weakening the effectiveness of the transition plane approach. 
To mitigate this issue, we propose an intra-to-inter face optimization strategy. %In the first stage, constraints are directly applied to the cube faces. In the second stage, the cube faces are stitched into a continuous panoramic image, followed by further fine-tuning to mitigate discontinuities in context resulting from panoramic stitching. 
First, we optimize 3D Gaussians with individual cube faces to enhance local details. After sufficient iterations, the rendered cube faces are stitched into an ERP panorama for content-consistent fine-tuning. This strategy effectively optimizes the 3D Gaussians while maintaining perceptual consistency across different faces. %Additionally, we introduce a spherical sampling technique, which involves padding the cube faces, effectively expanding overlapping regions, and thereby achieving seamless stitching across cube-face boundaries.
In addition, we introduce a spherical sampling technique to achieve seamless stitching across cube-face boundaries.

Extensive experiments evaluated on three popular real-world datasets—Rico360\cite{choi2023balanced}, OmniPhotos\cite{bertel2020omniphotos}, and OmniScene\cite{kim2021piccolo}—demons-trate that our proposed scheme outperforms the state-of-the-art approaches. Our contributions can be summarized as follows.
\begin{itemize}
    \item We propose \textbf{TPGS}, an effective framework to bridge continuous panoramic scene reconstruction with perspective Gaussian splatting.
    \item To address the challenges posed by the perspective cubemap projection, we propose a transition plane approach along with an intra-to-inter face optimization strategy to achieve effective optimization of 3D Gaussians in the panorama domain. 
    \item We introduce a spherical sampling approach to ensure seamless stitching across the boundaries of cube faces.
\end{itemize}

\begin{figure*}[t]
  \centering
  \includegraphics[width=0.95\textwidth]{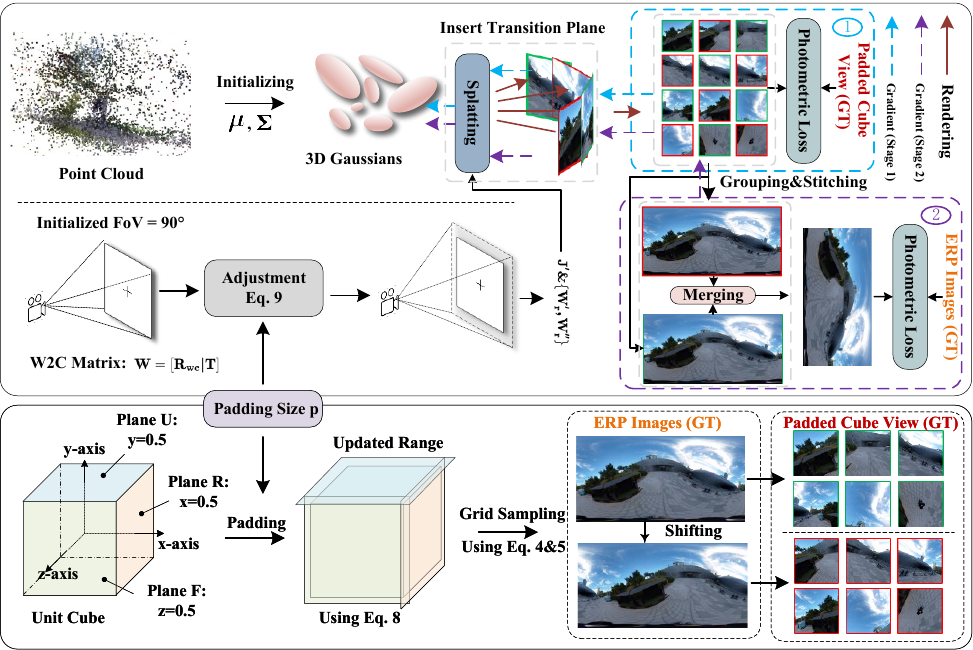} %1.png是图片文件的相对路径 
  \caption{The pipeline of the proposed TPGS.} %caption是图片的标题
  \label{fig:framework}
\end{figure*}
\section{Related Work}
\subsection{3D Scene Reconstruction}
3D scene reconstruction techniques\cite{huang2024360loc,huang2022360roam}, such as Structure-from-Motion (SfM)\cite{pagani2011structure,scaramuzza2006flexible,schonberger2016structure}, Multi-View Stereo (MVS)\cite{furukawa2015multi,habtegebrial2022somsi,meuleman2021real,sumikura2019openvslam}, and Simultaneous Localization and Mapping (SLAM)\cite{won2020omnislam,matsuki2024gaussian,sumikura2019openvslam}, recover point clouds and estimate camera poses by exploiting geometric consistency across views\cite{gava2023sphereglue}. Despite limitations in textureless or low-parallax regions, they remain effective for generating initial 3D geometry.
Neural Radiance Fields (NeRF)\cite{mildenhall2021nerf} introduced volumetric neural representations for high-quality novel view synthesis from sparse inputs. Many NeRF variants\cite{chen2022tensorf,gu2022omni,kulkarni2023360fusionnerf} aim to improve rendering speed \cite{li2022omnivoxel}, handle dynamic scenes\cite{wu20244d,yang2024deformable}, or incorporate geometric priors \cite{choi2023balanced,chen2023structnerf}. However, their high computational and memory costs hinder real-world deployment.
To address these issues, 3DGS has emerged as an efficient alternative. By leveraging differentiable Gaussian splats, 3DGS enables real-time, high-fidelity rendering and has become a strong baseline for novel view synthesis.
Recent work\cite{chen2024splatter,zhang2024pansplat} extends 3DGS to wide-angle and panoramic imagery to reduce input requirements. Especially, 360-GS \cite{bai2024360} improves indoor reconstruction by incorporating room layouts\cite{xu2021layout,rao2021omnilayout,shen2023disentangling,shen2024360} and planar priors \cite{sun2021indoor}, but its reliance on layout guidance limits generalization to more complex outdoor scenes. ODGS\cite{lee2024odgs} introduces a panorama-specific rasterizer and tangent-plane projection to bridge the panoramic–perspective domain gap. However, the local approximation introduces errors that require frequent Gaussian pruning and splitting during training, leading to instability and high memory use.
In contrast, our approach builds directly on the perspective 3DGS framework, enabling seamless integration of recent perspective-based advancements into panoramic reconstruction. This offers improved generality, efficiency, and scalability.

\subsection{Panoramic Distortion Processing}
Omnidirectional imaging has gained attention due to its ability to capture complete 360° scenes. However, the widely used equirectangular projection introduces significant distortions that hinder downstream 3D perception tasks\cite{liao2019dr, liao2024mowa}. To address panoramic distortions, prior works \cite{wang2020bifuse,shen2024revisiting,eder2019pano,liao2023cylin} have employed cubemap projection (or denser tangent-plane\cite{eder2020tangent} and icosahedral projections\cite{Ai_2024_CVPR}) for tasks like depth estimation, segmentation, and image synthesis. However, cubemap projection introduces projection distortion within each face and discontinuities across face boundaries. While often overlooked in other vision tasks, these limitations are particularly problematic in 3D Gaussian optimization, which requires strong cross-face and inter-face consistency. Alternative approaches, such as spherical CNNs \cite{coors2018spherenet,cohen2018spherical} and graph-based models \cite{yang2024360spred}, operate directly in the spherical domain but are incompatible with efficient rendering pipelines. Despite these efforts, panoramic distortion handling remains an open challenge in 3D reconstruction. Our approach addresses these challenges by introducing a transition plane to reduce cross-face and inner-face discontinuities and a spherical sampling strategy to alleviate stitching artifacts. Together, these improvements enable more accurate and consistent 3D reconstruction from omnidirectional images.

\section{Methodology}
\label{headings}
\subsection{Framework Overview}
%In this section, we introduce a novel panoramic 3D Gaussian Splatting framework designed for accurate and consistent 3D reconstruction from omnidirectional imagery. When panoramic images are converted to perspective views through cubemap projection, projection distortions and discontinuities at face boundaries occur, which pose challenges for consistent 3D reconstruction using 3DGS methods. To address these issues, our framework incorporates two core components: (1) a transition plane designed to generate intermediate views and explicitly handle discontinuities at face boundaries, thus effectively optimizing Gaussians in ambiguous boundary regions, and (2) a spherical sampling strategy that significantly alleviates boundary artifacts inherent in cubemap representations.

In this section, we introduce a novel panoramic 3D Gaussian Splatting framework designed for accurate and consistent 3D reconstruction from omnidirectional imagery. As illustrated in Fig.~\ref{fig:framework}, we first initialize the parameters (position, color, scale, rotation, and opacity) of the 3D Gaussians using a sparse point cloud generated by traditional Structure-from-Motion methods (\textit{e.g.,} OpenMVG\cite{moulon2017openmvg}). Subsequently, we employ a perspective camera-based rasterization pipeline, following the standard 3DGS approach, to render six individual cubemap faces. During rasterization, we employ a group of camera poses and the Jacobian matrices $J$ to adjust the splatting direction. This process generates intermediate views to explicitly address challenges introduced by the projection distortion, thus improving the stability of the Gaussian optimization.
In the optimization stage, we introduce an intra-to-inter face training strategy. Initially, the Gaussians are optimized primarily through photometric losses computed on individual cubemap faces and transition-plane views, enhancing local details. In the subsequent refinement stage, cross-face consistency constraints are implemented in the stitched ERP domain, promoting consistent representations across boundaries. By retaining compatibility with perspective-camera rasterization, our framework seamlessly integrates recent advances from perspective-based 3DGS research, enabling omnidirectional 3D reconstruction to benefit directly from ongoing developments in this rapidly evolving field.
\subsection{Transition Plane Gaussian Splatting}
In 3DGS, each 3D Gaussian is defined by a mean vector $\mu$ and a covariance matrix $\Sigma$. The covariance matrix is determined by the rotation matrix $\mathbf{R}$ and the scaling matrix $\mathbf{S}$, following the equation:
\begin{equation}
 \mathbf{\Sigma} = \mathbf{R}\mathbf{S}\mathbf{S}^T \mathbf{R}^T 
\end{equation}
Following the EWA splatting\cite{zwicker2002ewa} approach, the 3DGS \cite{matsuki2024gaussian} method approximates the 3D Gaussians as 2D Gaussians. The approximation can be simply represented as a transformation matrix $\mathbf{J}$. Although this transformation introduces some errors, it simplifies the projection process and reduces computational complexity. After approximation, the covariance of 2D Gaussians can be denoted as follows:
\begin{equation}
 \mathbf{\Sigma}_{2D} = \mathbf{J}\mathbf{W}\mathbf{\Sigma}\mathbf{W}^T \mathbf{J}^T 
\end{equation}
where $\mathbf{W}$ denotes the world-to-camera transformation matrix, it can represented as:
\begin{equation}
\mathbf{W} = [\mathbf{R_{wc}|T}] 
\end{equation}
Adjusting the rotation matrix $\mathbf{R_{wc}}$ in the world-to-camera transformation matrix allows us to modify the splatting orientation.

Our approach uses a rasterization pipeline for 3D Gaussians in a perspective camera. Thus, it can render only a perspective view with a limited FoV at a time. For subsequent 3D Gaussian optimization, we need to adjust the splatting orientation to align it with certain cube faces. To achieve this, we first review the cubemap projection. Specifically, we define a unit cube in Cartesian coordinates (shown in Figure \ref{fig:framework} bottom). The unit cube is centered at the origin with coordinate ranges from -0.5 to 0.5 on each axis. We generate a sampling grid for each face of the cube based on its geometric shape and position. For example, the front face is on the plane z=0.5 (back: z=-0.5, left: x=-0.5, right: x=0.5, up: y=0.5, down: y=-0.5), with the other two coordinates (x and y for front) sampled at equal intervals within [-0.5, 0.5], forming a regular sampling grid.

We then convert these 3D coordinates to spherical coordinates (longitude $\theta$ and latitude $\phi$), the procedure can be represented as follows:
\begin{eqnarray}
\left\{
\begin{aligned}
&\theta = \arctan(x, z)\\
&\phi= \arctan(y, \sqrt{x^2 + z^2})
\end{aligned}
\right.
\end{eqnarray}
Finally, we convert the spherical coordinates to pixel coordinates in the equirectangular projection using:
\begin{eqnarray}
\left\{
\begin{aligned}
u = \left(\frac{\theta}{2\pi} + 0.5\right) \times \text{W} - 0.5\\
v = \left(-\frac{\phi}{\pi} + 0.5\right) \times \text{H} - 0.5
\end{aligned}
\right.
\end{eqnarray}
\noindent \textbf{Transition Plane}. As shown in Eq. 4, the conversion from 3D coordinates to spherical coordinates relies on the arctan function. The nonlinear nature of this function leads to uneven sampling density in the cube map projection, with higher density in the center and lower density at the edges. Hence, the boundary regions of cube faces typically suffer from sparse sampling and significant distortion. To address this issue, we propose adding transition planes at the boundaries between adjacent cube faces. The center of these planes lies at the intersection of adjacent faces, with their edges complemented by the central areas of the neighboring faces. Consequently, the transition planes provide a more uniform sampling density, effectively alleviating distortion in the cube map projection. We implement the transition plane in detail as follows.

Based on the cubemap projection transformation definition, we can conveniently change the splatting direction by updating the world-to-camera transformation matrix. Specifically, during the rendering phase, we assume each cubemap face has a 90° field of view. Under this setup, directly rendering with the initial world-to-camera transformation matrix $\mathbf{W}$ yields the front view of the cube. To obtain other views, we simply define two types of rotation matrices: 
\[
\mathbf{R}_x(\psi) = \begin{bmatrix}
1 & 0 & 0 \\
0 & \cos\psi & -\sin\psi \\
0 & \sin\psi & \cos\psi
\end{bmatrix}
\]
\[
\mathbf{R}_y(\psi) = \begin{bmatrix}
\cos\psi & 0 & \sin\psi \\
0 & 1 & 0 \\
-\sin\psi & 0 & \cos\psi
\end{bmatrix}
\]
where $\mathbf{R}_x(\psi)/\mathbf{R}_y(\psi)$ denotes that rotates $\psi$ around x/y axis, respectively. $\mathbf{R}_y(\psi)$ is employed to get other horizontal faces (left, right, back), while $\mathbf{R}_x(\psi)$ vertical ones (up, down). Therefore, the updated world-to-camera transformation matrix can be written as:
\begin{equation}
\label{e6}
    \mathbf{W}_r = \left\{
\begin{aligned}
&[\mathbf{R_{wc}}\mathbf{R}_y^T(\psi)|\mathbf{T}],\quad \psi \in \{0^\circ, 90^\circ, 180^\circ, 270^\circ\}\\
&[\mathbf{R_{wc}}\mathbf{R}_x^T(\psi)|\mathbf{T}],\quad \psi \in \{-90^\circ, 90^\circ\}
\end{aligned}
\right.
\end{equation}
Therefore, we can employ $\mathbf{W}_r$ to get the original six cube faces.
To alleviate the projection distortion introduced by cubemap projection, we introduce a transition plane. This method optimizes ambiguous-region 3D Gaussians by rendering intermediate perspective views during rasterization. The intermediate perspective can be achieved by rotating 45° around the y-axis. The updated rotation matrix can be expressed as:
\begin{eqnarray}
\mathbf{R_{wc}}' = \mathbf{R_{wc}}\mathbf{R}_y^T(45^\circ)
\end{eqnarray}
By applying Eq. \ref{e6}, we can obtain another group of updated world-to-camera transformation matrices $\mathbf{W}_r'$ associated with the transition plane. Using $\mathbf{W}_r'$ and $\mathbf{W}_r$, we render 12 views—six corresponding to the original cube faces and six intermediate views associated with the transition planes—for subsequent optimization.
\begin{table*}[t]
\centering
\caption{Quantitative comparison of 3D reconstruction results on OmniScenes dataset. The best result for each metric is written in bold. Each cell contains PSNR, SSIM, and LPIPS in sequence.}
\label{tab:omniscenes}
\scalebox{0.83}{\begin{tabular}{c|c|c|c|c|c|c}
\hline
Scene & NeRF(P) & 3DGS(P) & TensoRF & EgoNeRF & ODGS& Ours\\ \hline
pyebachRoom 1 & 14.76 / 0.5940 / 0.5417 & 20.56 / 0.7347 / 0.2525 & 21.69 / 0.6800 / 0.4751 & 21.13 / 0.7055 / 0.3934 &\textbf{ 22.78} / 0.8038 / \textbf{0.1449}&22.74/ \textbf{0.8103}/ 0.1614 \\ 

 room 1 & 15.14 / 0.7264 / 0.4518 & 22.48 / 0.8566 / 0.1985 & 19.81 / 0.8159 / 0.3020 & 21.68 / 0.7908 / 0.3358 &  -/-/-&\textbf{23.80/ 0.8647/ 0.0997} \\ 
 
 room 2 & 15.09 / 0.7156 / 0.4563 & 22.90 / 0.8277 / 0.1979 & 23.24 / 0.8042 / 0.3425 & 22.51 / 0.7837 / 0.3411 & \textbf{24.17} / 0.8303 / 0.1302&24.02/ \textbf{0.8401/ 0.1294} \\ 
 
 room 3 & 16.21 / 0.7811 / 0.3790 & 25.13 / 0.8860 / 0.1554 & 26.39 / 0.8708 / 0.3022 & 22.79 / 0.8423 / 0.3454 & 24.08 / 0.8743 / 0.1384&\textbf{26.22/ 0.9016/ 0.1256} \\
 
 room 4 & 15.52 / 0.7467 / 0.4202 & 25.61 / 0.8816 / 0.1656 & 24.97 / 0.8585 / 0.2901 & 24.66 / 0.8434 / 0.3117 & 26.14 / 0.8960 / \textbf{0.1052}&\textbf{26.55/ 0.8997}/ 0.1171 \\   
 
 room 5 & 16.75 / 0.8105 / 0.3805 & 24.43 / 0.8185 / 0.1843 & 25.52 / 0.8855 / 0.3198 & 22.87 / 0.8585 / 0.3416 & 24.49 / 0.8819 / 0.1493&\textbf{25.66/ 0.9009/ 0.1396} \\ 
 
 weddingHall 1 & 16.40 / 0.6783 / 0.5592 & 24.18 / 0.8326 / 0.8304 & 23.42 / 0.7680 / 0.4352 & 23.85 / 0.7741 / 0.3552 & 24.83 / 0.8347 / \textbf{0.1664}&\textbf{25.18/ 0.8440}/ 0.1807 \\ \hline
 
 Average & 15.79 / 0.7210 / 0.4551 & 23.80 / 0.8424 / 0.2977 & 24.21 / 0.8111 / 0.3619 & 22.97 / 0.8012 / 0.3481 & 24.42 / 0.8526 / 0.1391&\textbf{24.88/ 0.8659/ 0.1362} \\ \hline \hline
%%%%%%%%%%%%%%%%%%%%%%%%%%%%%%%%%%%%%%%%%%%%%%%%%%
pyebachRoom 1 & 15.99 / 0.5932 / 0.5669 & 17.71 / 0.6282 / 0.4094 & 22.57 / 0.6993 / 0.4026 & \textbf{23.69} / 0.7688 / 0.2787& 22.91 / 0.8068 / \textbf{0.1322} &23.61/ \textbf{0.8187}/ 0.1334  \\ 

 room 1 & 14.64 / 0.6604 / 0.4850 & 17.14 / 0.5864 / 0.5746 & 19.99 / 0.8226 / 0.2797 & 22.44 / 0.8083 / 0.2715 & -/-/-&23.89/ 0.8577/ 0.1009 \\ 
 
 room 2 & 14.66 / 0.6464 / 0.5138 & 11.68 / 0.6808 / 0.5027 & 23.58 / 0.8083 / 0.3037 & 23.53 / 0.8086 / 0.4122 & 23.13 / 0.8108 / \textbf{0.1422}&\textbf{23.58/ 0.8169}/ 0.1464\\ 
 
 room 3 & 16.37 / 0.6757 / 0.4280 & 20.13 / 0.8066 / 0.3109 & 27.26 / 0.8793 / 0.2631 & 26.43 / 0.8789 / 0.2364 & 25.56 / 0.8847 / 0.1150&\textbf{26.56/ 0.8993/ 0.1106} \\ 
 
 room 4 & 16.13 / 0.6713 / 0.4724 & 19.18 / 0.7712 / 0.3390 & 26.06 / 0.8682 / 0.2495 & 26.17 / 0.8693 / 0.2336 & 26.25 / 0.8879 / \textbf{0.1025}& \textbf{26.82/ 0.8964}/ 0.1072\\ 
 
 room 5 & 17.43 / 0.7996 / 0.4195 & 17.93 / 0.8088 / 0.3336 & 26.01 / 0.8894 / 0.2834 & \textbf{26.06} / 0.8803 / 0.2652 & 24.44 / 0.8797 / 0.1383&25.98/ \textbf{0.9005/ 0.1230} \\ 
 
 weddingHall 1 & 16.64 / 0.6398 / 0.5541 & 21.25 / 0.7818 / 0.2642 & 24.04 / 0.7786 / 0.3818 & 25.00 / 0.8052 / 0.2718 & 24.77 / 0.8331 / \textbf{0.1392}&\textbf{25.49/ 0.8404}/ 0.1412 \\ \hline
 
 Average & 16.20 / 0.6838 / 0.4925 & 18.08 / 0.7329 / 0.3600 & 24.92 / 0.8205 / 0.3140 & \textbf{25.15} / 0.8352 / 0.2607 & 24.51 / 0.8505 / 0.1282&25.13/ \textbf{0.8614/ 0.1232}  \\ \hline
\end{tabular}}
\end{table*}
\subsection{Cube Padding with Spherical Sampling}
As previously mentioned, optimizing ambiguous-region 3D Gaussians (3DGS) can be achieved through rendering based on transition planes. However, synthesizing panoramic images from novel viewpoints still requires inverse transformation from perspective projections (obtained via 3DGS rendering) back into the ERP format. During this inverse transformation, sampling errors due to projection can introduce noticeable stitching artifacts in the resulting panoramic images.

To address this issue, we propose a cubemap projection padding method based on spherical sampling. Specifically, our padding strategy leverages the unit cube representation introduced in Section 3.1 by adjusting the sampling grid range. Suppose the input ERP image has a resolution of \(H\times W \), and we aim to obtain a cube face with padding size \(p\) (in pixels); then, the updated sampling grid range becomes:
\begin{equation}
[-0.5 - \frac{2p}{H},\quad 0.5 + \frac{2p}{H}]    
\end{equation}
Simultaneously, we update the FoV in 3DGS to render padded perspective views for alignment. The procedure can be denoted as follows:
\begin{equation}
fov' = \frac{2\pi p}{H}+ \frac{\pi}{2}   
\end{equation}
With this simple adjustment, effective padding for cube faces can be conveniently achieved. This idea shares a similar motivation with the spherical padding method (e.g., Bifuse \cite{wang2020bifuse}), but our approach is simpler in implementation and avoids potential negative impacts on 3DGS optimization caused by the introduction of additional camera intrinsic.

\begin{figure}[t]
  \centering
  \includegraphics[width=0.4\textwidth]{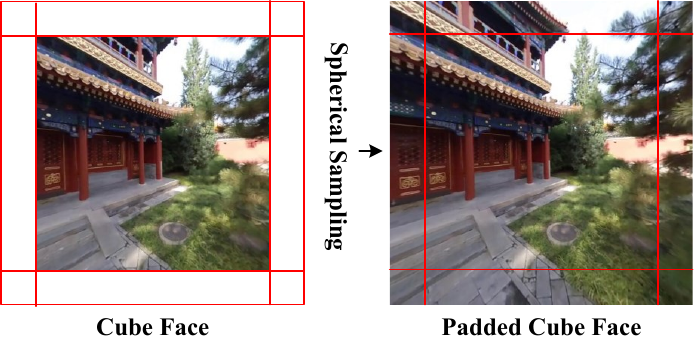} %1.png是图片文件的相对路径 
  \caption{Cube padding with spherical sampling.} %caption是图片的标题
  \label{fig:pad}
\end{figure}
\subsection{Intra-to-Inter Face Optimization}
We strictly follow previous works in using photometric loss to optimize the 3D Gaussians. The objective function consists of two terms: an \(\mathcal{L}_1\) term and an \(\mathcal{L}_{\text{D-SSIM}}\) term. The former imposes a pixel-wise constraint, while the latter is suited for capturing structural similarity.

Firstly, we focus on intra-face optimization by constructing the objective function using rendered perspective views. After each gradient backpropagation, we perform pruning and splitting operations on the 3D Gaussians to ensure that a sufficient number of points are accumulated early in the optimization. This intra-face optimization ensures that the 3D Gaussian parameters are optimized for each face, aligning well with the specific rendered views and optimizing them for pixel-wise accuracy. The objective function in this stage is formulated as:
\begin{equation}
\mathcal{L} = \lambda_1 \mathcal{L}_1(V_r, V_{gt}) + \lambda_2 \mathcal{L}_{\text{D-SSIM}}(V_r, V_{gt})
\end{equation}
where \( V_r \) and \( V_{gt} \) denote the rendered perspective views and the corresponding ground-truth images obtained via cube projection, respectively.

Secondly, we perform inter-face optimization by optimizing the consistency between different cube faces through stitching perspective views into ERP panorama images. Let \( E_c \) denote the ERP image reconstructed from cube faces, and \( E_t \) denote the ERP image derived from transition planes. To align \( E_t \) with \( E_c \), we inversely rotate \( E_t \) by 45°. The final rendered panoramic image \( E_r \) is obtained by averaging \( E_c \) and \( E_t \), formulated as:
\begin{equation}
E_r = \frac{\text{Rotate}(E_t, -45^\circ) + E_c}{2}
\end{equation}
In this stage, inter-face optimization is achieved by aligning the global panoramic image with the ground-truth ERP image, optimizing over the global structure and consistency between different cube faces. The objective function for the second stage is defined as:
\begin{equation}
\mathcal{L} = \lambda_1 \mathcal{L}_1(E_r, E_{gt}) + \lambda_2 \mathcal{L}_{\text{D-SSIM}}(E_r, E_{gt}) 
\end{equation}
where \( E_{gt} \) represents the ground-truth panoramic image, and \( \lambda_1, \lambda_2 \) are balancing factors. We strictly follow the previous approach\cite{matsuki2024gaussian,lee2024odgs} by setting \( \lambda_1 = 0.8 \) and \( \lambda_2 = 0.2 \).
\begin{table*}[t]
\centering
\caption{Quantitative comparison of 3D reconstruction results on Ricoh360 dataset. The best result for each metric is written in bold. Each cell contains PSNR, SSIM, and LPIPS in sequence.}
\label{tab:rico360}
\scalebox{0.85}{\begin{tabular}{c|c|c|c|c|c|c}
\hline
Scene & NeRF(P) & 3DGS(P) & TensoRF & EgoNeRF & ODGS &Ours \\
%%%%%%%%%%%%%10min%%%%%%%%%%%%%%%%%%%%%%%%%%
\hline
 bricks & 12.40 / 0.4489 / 0.6403 & 23.88 / 0.7875 / 0.2275 & 21.08 / 0.6201 / 0.4918 & 23.09 / 0.7223 / 0.2984 & 23.90 / 0.8185 / \textbf{0.1296}& \textbf{24.61}/ \textbf{0.8353}/ 0.1436 \\
 
 bridge & 14.96 / 0.5553 / 0.5866 & 23.78 / 0.7783 / 0.2145 & 21.93 / 0.6456 / 0.4823 & 23.34 / 0.7199 / 0.3096 & 23.88 / 0.7987 /\textbf{ 0.1286}&\textbf{24.36}/ \textbf{0.8101}/ 0.1507 \\
 
 bridge under & 19.71 / 0.4987 / 0.6769 & 24.30 / 0.7892 / 0.2239 & 21.99 / 0.6323 / 0.5791 & 24.11 / 0.7504 / 0.3202 & 25.12 / 0.8347 / \textbf{0.1339} &\textbf{26.09}/ \textbf{0.8583}/ 0.1379 \\
 
 cat tower & 12.54 / 0.5182 / 0.5990 & 24.33 / 0.7543 / 0.2548 & 22.45 / 0.6308 / 0.6002 & 23.80 / 0.6861 / 0.3758 & 24.47 / 0.7771 / \textbf{0.1435}&\textbf{25.14}/ \textbf{0.7900}/ 0.1882 \\
 
 center & 14.76 / 0.6691 / 0.5211 & 27.24 / 0.8364 / 0.2887 & 27.23 / 0.8088 / 0.4294 & 27.97 / 0.8450 / 0.2521 & 28.10 / 0.8710 / 0.1206&\textbf{29.10}/ \textbf{0.8957}/ \textbf{0.1041} \\
 
 farm & 14.45 / 0.4970 / 0.6262 & 21.66 / 0.6897 / 0.3248 & 20.80 / 0.5683 / 0.5141 & 21.80 / 0.6483 / 0.3386 & 20.74 / 0.6881 / 0.2270&\textbf{22.15}/ \textbf{0.7400}/ \textbf{0.2199} \\
 
 flower & 12.03 / 0.4132 / 0.6912 & 21.71 / 0.6942 / 0.3247 & 20.07 / 0.5414 / 0.6696 & 21.51 / 0.6149 / 0.4211 & 22.19 / 0.7273 / \textbf{0.1925}&\textbf{22.67}/ \textbf{0.7386}/ 0.2407 \\
 
 gallery chair & 15.40 / 0.6950 / 0.4929 & 27.76 / 0.8732 / 0.1962 & 26.00 / 0.7907 / 0.5233 & 27.01 / 0.8326 / 0.3409 & 27.29 / 0.8777 / 0.1353&\textbf{28.22}/ \textbf{0.8940}/ \textbf{0.1324} \\
 
 gallery park & 12.29 / 0.6050 / 0.5176 & 25.30 / 0.8021 / 0.2384 & 24.21 / 0.7394 / 0.5120 & 25.11 / 0.7703 / 0.3245 & 25.48 / 0.8241 / \textbf{0.1341}&\textbf{26.34}/ \textbf{0.8349}/ 0.1505 \\
 
 gallery pillar & 14.50 / 0.6445 / 0.4902 & 27.79 / 0.8613 / 0.1617 & 25.85 / 0.7821 / 0.3977 & 27.31 / 0.8312 / 0.2379 & 28.02 / 0.8821 / \textbf{0.0882}&\textbf{28.69}/ \textbf{0.8914}/ 0.0976 \\
 
 garden & 13.97 / 0.5682 / 0.5430 & 27.53 / 0.7919 / 0.2118 & 25.37 / 0.6649 / 0.5616 & 26.48 / 0.7175 / 0.3517 & 23.20 / 0.7843 / 0.2289&\textbf{27.92}/ \textbf{0.8219}/ \textbf{0.1634} \\
 
 poster & 14.99 / 0.6258 / 0.5679 & 26.14 / 0.8599 / 0.2098 & 23.20 / 0.7500 / 0.4784 & 25.39 / 0.8213 / 0.3205 & 26.90 / 0.8782 / \textbf{0.1249}&\textbf{27.47}/ \textbf{0.8934}/ 0.1268 \\
\hline
 Average & 14.33 / 0.5616 / 0.5794 & 25.12 / 0.7932 / 0.2397 & 23.35 / 0.6812 / 0.5200 & 24.74 / 0.7467 / 0.3243 & 24.94 / 0.8135 / \textbf{0.1489} & \textbf{26.40}/ \textbf{0.8532}/ 0.1581\\
\hline \hline

%%%%%%%%%%%%%100min%%%%%%%%%%%%%%%%%%%%%%%%%%
 bricks & 15.01 / 0.4760 / 0.6245 & 22.60 / 0.7410 / 0.2855 & 21.66 / 0.6353 / 0.4375 & 23.93 / 0.7616 / 0.2475 & 24.62 / 0.8479 / \textbf{0.1021} & \textbf{25.32}/ \textbf{0.8563}/ 0.1111\\

 bridge & 17.32 / 0.5558 / 0.5620 & 21.94 / 0.7157 / 0.3133 & 22.58 / 0.6558 / 0.4306 & 23.94 / 0.7516 / 0.2562 & 24.37 / 0.8154 / \textbf{0.1063}&\textbf{24.86}/ \textbf{0.8274}/ 0.1167 \\

 bridge under & 16.42 / 0.5075 / 0.6447 & 19.03 / 0.6377 / 0.3663 & 22.86 / 0.6577 / 0.4826 & 25.05 / 0.7924 / 0.2492 & 25.93 / 0.8538 / 0.1026&\textbf{27.03}/ \textbf{0.8808}/ \textbf{0.0953}\\

 cat tower & 15.45 / 0.5323 / 0.5824 & 21.24 / 0.6851 / 0.3565 & 23.02 / 0.6393 / 0.5477 & 25.05 / 0.7163 / 0.3417 & 25.35 / \textbf{0.8088} / \textbf{0.1109}&\textbf{25.48}/ 0.8042/ 0.1353 \\

 center & 17.09 / 0.6566 / 0.4955 & 20.04 / 0.6974 / 0.4237 & 27.90 / 0.8182 / 0.3840 & 29.12 / 0.8625 / 0.2119 & 29.39 / 0.8940 / \textbf{0.0808}&\textbf{29.72}/ \textbf{0.9034}/ 0.0827 \\

 farm & 15.93 / 0.4830 / 0.6173 & 21.49 / 0.6844 / 0.3299 & 21.09 / 0.5765 / 0.4662 & 22.25 / 0.6745 / 0.3089 & 16.34 / 0.6039 / 0.4109&\textbf{22.40}/\textbf{ 0.7601}/ \textbf{0.1550} \\

 flower & 13.57 / 0.4153 / 0.6845 & 20.48 / 0.6559 / 0.3531 & 20.57 / 0.5506 / 0.6161 & 22.08 / 0.6497 / 0.3922 & 22.71 / 0.7485 / \textbf{0.1509}&\textbf{22.93}/ \textbf{0.7554}/ 0.1703 \\

 gallery chair & 17.59 / 0.6873 / 0.5240 & 26.44 / 0.8509 / 0.2161 & 26.61 / 0.7990 / 0.4766 & 27.71 / 0.8505 / 0.2993 & 27.62 / 0.8831 / 0.1135&\textbf{28.76}/ \textbf{0.9035}/ \textbf{0.1059} \\

 gallery park & 14.24 / 0.5847 / 0.5404 & 23.22 / 0.7637 / 0.3027 & 24.64 / 0.7457 / 0.4724 & 25.64 / 0.7848 / 0.3001 & 26.19 / 0.8401 / \textbf{0.1076}&\textbf{26.83}/ \textbf{0.8459}/ 0.1172 \\

 gallery pillar & 16.57 / 0.6405 / 0.4946 & 21.93 / 0.7429 / 0.3205 & 26.49 / 0.7960 / 0.3336 & 27.97 / 0.8467 / 0.2104 & 28.74 / 0.8970 / \textbf{0.0693}&\textbf{29.20}/ \textbf{0.8994}/ 0.0816 \\

 garden & 17.81 / 0.5840 / 0.5101 & 25.97 / 0.7792 / 0.2528 & 25.91 / 0.6738 / 0.5201 & 27.16 / 0.7441 / 0.3112 & 27.09 / \textbf{0.8383} / \textbf{0.1006}&\textbf{28.34}/ 0.8364/ 0.1178 \\

 poster & 16.91 / 0.6169 / 0.5794 & 20.45 / 0.7199 / 0.3406 & 24.32 / 0.7750 / 0.4161 & 26.50 / 0.8497 / 0.2613 & 26.92 / 0.8808 / 0.1113&\textbf{28.28}/ \textbf{0.9008}/ \textbf{0.1077} \\
\hline

 Average & 16.16 / 0.5617 / 0.5716 & 22.07 / 0.7228 / 0.3218 & 23.97 / 0.6936 / 0.4653 & 25.49 / 0.7737 / 0.2825 & 25.44 / 0.8260 / 0.1306&\textbf{26.94}/ \textbf{0.8554}/ \textbf{0.1305} \\
\hline
\end{tabular}}
\end{table*}

\begin{table*}[t]
\centering
\caption{Quantitative comparison of 3D reconstruction results on OmniPhotos dataset. The best result for each metric is written in bold. Each cell contains PSNR, SSIM, and LPIPS in sequence.}
\label{tab:omniphoto}
\scalebox{0.83}{\begin{tabular}{c|c|c|c|c|c|c}
\hline
Scene & NeRF(P) & 3DGS(P) & TensoRF & EgoNeRF & ODGS& Ours\\ \hline
Ballintoy & 19.90 / 0.7292 / 0.4717 & 28.67 / 0.8875 / 0.2094 & 25.68 / 0.8008 / 0.4190 & 28.49 / 0.8715 / 0.2270 & 29.11 / 0.9085 / \textbf{0.1076}&\textbf{29.95/ 0.9157}/ 0.1119 \\ 

BeihaiPark & 16.76 / 0.5946 / 0.5871 & 23.99 / 0.8126 / 0.2600 & 22.16 / 0.6855 / 0.5516 & 24.35 / 0.7755 / 0.2682 & 25.34 / 0.8600 / 0.1409&\textbf{26.05/ 0.8848/ 0.1193} \\ 

Cathedral & 15.84 / 0.4736 / 0.6851 & 23.01 / 0.7885 / 0.2569 & 19.35 / 0.6368 / 0.5511 & 23.11 / 0.7267 / 0.2898 & 23.74 / 0.8394 / \textbf{0.1416}&\textbf{24.37/ 0.8497}/ 0.1517 \\

Coast & 20.38 / 0.6452 / 0.5236 & 27.86 / 0.8378 / 0.2011 & 24.69 / 0.7026 / 0.4440 & 27.74 / 0.8006 / 0.2320 & 28.75 / 0.8837 / \textbf{0.0966}&\textbf{29.31/ 0.8908}/ 0.1030 \\ 

Field & 23.44 / 0.7063 / 0.4293 & 29.51 / 0.8392 / 0.1600 & 27.25 / 0.7427 / 0.4602 & 28.53 / 0.7843 / 0.2618 & 29.71 / 0.8702 / \textbf{0.0892}&\textbf{30.25/ 0.8738}/ 0.1082 \\ 

Nunobiki2 & 18.79 / 0.6182 / 0.5295 & 24.84 / 0.8017 / 0.2224 & 22.93 / 0.6719 / 0.5308 & 23.45 / 0.7052 / 0.3378 & 21.62 / 0.8289 / 0.1496&\textbf{25.50/ 0.8545/ 0.1392} \\

SecretGarden1 & 16.70 / 0.6380 / 0.5040 & 24.58 / 0.8579 / 0.1706 & 22.49 / 0.7066 / 0.5205 & 24.87 / 0.7885 / 0.2450 &26.33 / 0.8786 / 0.1023 &\textbf{27.12/ 0.8993/ 0.0864} \\

Shrines1 & 15.67 / 0.4516 / 0.7258 & 22.36 / 0.7449 / 0.2864 & 19.52 / 0.5179 / 0.6441 & 21.28 / 0.6374 / 0.3693 & \textbf{23.45 / 0.8108 / 0.1560}&23.31/ 0.8077/ 0.1821 \\

Temple3 & 21.17 / 0.5932 / 0.6112 & 25.17 / 0.8549 / 0.1839 & 20.83 / 0.6000 / 0.5691 & 24.31 / 0.7916 / 0.2321 & 26.14 / 0.8881 / \textbf{0.0868}&\textbf{26.50/ 0.8962}/ 0.0881 \\

Wulongting & 17.89 / 0.7083 / 0.4451 & 25.80 / 0.8845 / 0.1497 & 22.89 / 0.7697 / 0.3987 & 25.89 / 0.8405 / 0.1991 & 26.98 / 0.9203 / \textbf{0.0626}&\textbf{27.43/ 0.9240}/0.0684 \\ \hline

Average & 18.14 / 0.6158 / 0.5514 & 25.61 / 0.8310 / 0.2100 & 22.78 / 0.6841 / 0.5089 & 25.20 / 0.7722 / 0.2662 & 26.24 / 0.8704 /\textbf{ 0.1108}& \textbf{26.98/ 0.8777}/ 0.1158 \\ \hline \hline
%%%%%%%%%%%%%%%%%%%%%%%%%%%%%%%%%%%%%%%%%%%%%%%%%%
Ballintoy & 23.32 / 0.7358 / 0.3796 & 26.37 / 0.8694 / 0.2226 & 26.85 / 0.8193 / 0.3738 & 29.23 / 0.8981 / 0.1827 & 30.06 / 0.9220 / \textbf{0.0780}&\textbf{30.84/ 0.9235}/ 0.0873 \\

 BeihaiPark & 18.48 / 0.6089 / 0.4949 & 20.81 / 0.7581 / 0.2898 & 23.13 / 0.7095 / 0.4783 & 26.19 / 0.8390 / 0.1778 & 19.94 / 0.7955 / 0.2352&\textbf{26.77/ 0.8953/ 0.0904} \\ 
 
 Cathedral & 18.01 / 0.5032 / 0.6033 & 21.20 / 0.6879 / 0.3682 & 20.23 / 0.5935 / 0.4848 & 24.88 / 0.8056 / 0.1848 & \textbf{25.39 / 0.8769 / 0.1081}&25.18/ 0.8668/ 0.1198 \\
 
 Coast & 23.12 / 0.6666 / 0.4263 & 27.29 / 0.8334 / 0.2091 & 26.27 / 0.7392 / 0.3868 & 29.28 / 0.8553 / 0.1627 & 29.23 / 0.9004 / 0.0770&\textbf{30.02/ 0.9043/ 0.0770} \\  
 
Field & 25.81 / 0.7274 / 0.3980 & 29.36 / 0.8420 / 0.1665 & 27.72 / 0.7488 / 0.4209 & 30.04 / 0.8774 / 0.1776 & 29.18 / 0.8866 / 0.0851& \textbf{31.17/ 0.8902/ 0.0763} \\ 

Nunobiki2 & 21.02 / 0.6346 / 0.4705 & 20.30 / 0.7116 / 0.3662 & 23.63 / 0.6666 / 0.4686 & 25.13 / 0.7879 / 0.2115 & 25.78 / \textbf{0.8659 / 0.0996}&\textbf{26.03}/ 0.8645/ 0.1091 \\  

SecretGarden1 & 16.70 / 0.6380 / 0.5040 & 22.49 / 0.7066 / 0.5205 & 24.87 / 0.7885 / 0.2450 & 27.53 / 0.8940 / 0.0769 & 27.53 / 0.8940 / 0.0769&\textbf{27.82/ 0.9040/ 0.0682} \\ 

Shrines1 & 18.13 / 0.4803 / 0.6417 & 23.93 / 0.8470 / 0.2015 & 23.39 / 0.7273 / 0.4548 & 26.76 / 0.8514 / 0.1513 & 23.94 / \textbf{0.8270 / 0.1319}&\textbf{23.95} / 0.8226 / 0.1419 \\  

Temple3 & 18.69 / 0.6232 / 0.5202 & 22.22 / 0.7889 / 0.2523 & 24.02 / 0.7875 / 0.4344 & 26.21 / 0.8914 / 0.1411 & 24.58 / 0.8895 / 0.0822&\textbf{27.24/ 0.9025/ 0.0734}  \\ 

Wulongting & 20.98 / 0.7328 / 0.3736& 21.23 / 0.7945 / 0.2964& 24.02 / 0.7875 / 0.3434& 27.84 / 0.8914 / 0.1240& 27.66 / 0.9255 / \textbf{0.0531}&\textbf{28.38/ 0.9295}/ 0.0560 \\ \hline

Average & 20.80 / 0.6388 / 0.4772 & 23.30 / 0.7859 / 0.2670 & 23.73 / 0.7038 / 0.3344 & 26.90 / 0.8349 / 0.1766 & 26.33 / 0.8786 / 0.1023&\textbf{27.74/ 0.8903/ 0.0899} \\ \hline

\end{tabular}}
\end{table*}
\begin{figure*}[t]
  \centering
  \includegraphics[width=\textwidth]{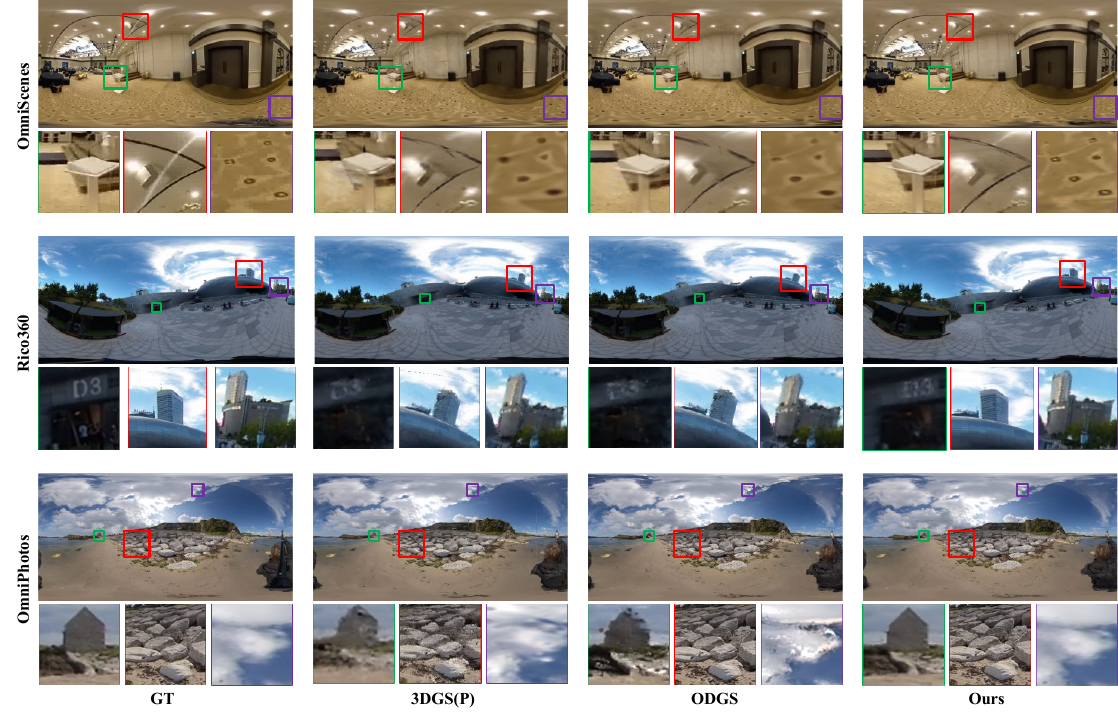} %1.png是图片文件的相对路径 
  \caption{Qualitative comparison results.} %caption是图片的标题
  \label{fig:quali}
\end{figure*}
\section{Experiment}
In this study, we use three representative real-world\footnote{The adjusted version of the synthetic datasets (360Roam\cite{huang2022360roam}, OmniBlender\cite{choi2023balanced}, and 360VO\cite{huang2022360vo}) is not released.} datasets, each covering different scenes (including egocentric, roaming, indoor, and outdoor scenes), to validate the universality and effectiveness of our method.
\subsection{Datasets}
The Ricoh360 dataset contains 12 omnidirectional outdoor scenes, each captured by rotating in place at a fixed location. Given the in-place rotation method, the images maintain strong continuity and consistency in their perspectives, making this dataset ideal for testing the method's performance in static environments. The OmniPhotos dataset presents a different shooting method, comprising 10 omnidirectional scenes captured using a 360° camera mounted on a selfie stick with a circular motion. The diversity of this dataset, with variations in shooting paths, makes it well-suited to test the method’s performance in dynamic environments. The OmniScenes dataset includes 7 indoor scenes, originally designed to assess the visual localization quality of omnidirectional images in harsh conditions. This dataset features challenging factors such as motion blur, scene changes, and image compression artifacts, making it ideal for testing the robustness of the algorithm under imperfect data conditions. These datasets encompass different shooting modes and environmental conditions, ensuring that our method is thoroughly validated across various real-world scenarios. The differences between the datasets further ensure the generalizability of the results, allowing us to evaluate the method’s performance comprehensively. We strictly follow the previous work\cite{lee2024odgs} to split the training and testing sets.

\subsection{Metrics}
Following the previous works\cite{bai2024360,lee2024odgs}, we evaluate the quality of 3D reconstructions using three widely adopted image quality metrics: Peak Signal-to-Noise Ratio (PSNR), Structural Similarity Index\cite{wang2004image} (SSIM) and Learned Perceptual Image Patch Similarity\cite{zhang2018unreasonable} (LPIPS). %PSNR is a traditional metric that measures the overall similarity between two images based on pixel differences. SSIM evaluates the perceptual quality of images by considering luminance, contrast, and structural information. It aligns more closely with human visual perception by considering structural aspects of the images, making it more sensitive to perceptual changes in texture and structure compared to PSNR. LPIPS  is a deep learning-based metric that evaluates perceptual differences between images by comparing high-level feature representations obtained from pre-trained deep neural networks (basically AlexNet in this paper). %
These metrics are selected to assess different aspects of image quality: PSNR evaluates brightness fidelity, SSIM measures structural similarity, and LPIPS assesses perceptual similarity. Using these metrics ensures a comprehensive evaluation from pixel-level, structural, and perceptual perspectives.
\subsection{Comparison Results}
\noindent \textbf{Quantitative comparison.} %We strictly follow the experimental setting of ODGS for evaluation, that is, we evaluate the performance of all methods within 10 and 100 minutes of training time. We compare the 3DGS method based on perspective projection (referred to as 3DGS(P) in the table), the NeRF method (NeRF(P), TensorNeRF) and the panoramic customization method (EgoNeRF, ODGS). Tables 1, 2, and 3 show the performance of our method and the state-of-the-art methods on the three datasets. As can be seen from the table, our method outperforms other methods in almost all scenes, especially in the SSIM indicator. This shows that the new perspective rendered by our method is more in line with the human eye perception characteristics and is conducive to the immersive experience of the scene. We observe that NeRF and TensoRF based on Cartesian coordinate grids perform poorly in representing large outdoor scenes and lag behind other methods significantly. As can be seen from the table, the EgoNeRF method with a spherical balanced grid can obtain better PSNR and SSIM, but it still lags behind the state-of-the-art methods significantly in the LPIPS indicator. In addition, such NeRF-based methods require longer rendering times, which poses a challenge to real-time requirements. 3DGS with perspective images suffers from severe overfitting. Since there is no overlap between the six faces after cubemap decomposition, 3DGS performs six independent optimizations for faces facing the same direction. Therefore, even if the input is the same, the amount of information used is significantly reduced, resulting in overfitting that occurs quickly. The panoramic customized ODGS method shows suboptimal performance except for ours. Since the spherical cut approximation reintroduces projection errors, a large number of redundant 3D Gaussians will be generated in large distortion areas, which will significantly increase the demand for video memory. In addition, this error will cause unstable training and significantly affect the performance in some scenes. As shown in Table 2, ODGS fails on the room1 scene of the OmniScene dataset, so no results are reported. Despite this, our method still achieves the best performance. In terms of rendering time, ODGS claims that their method reaches real-time. However, our method can theoretically render in parallel to reduce training time and achieve comparable rendering speed. \\
We strictly follow the experimental setting of ODGS\cite{lee2024odgs} for evaluation. Specifically, we evaluate the performance of all methods within 10 (the first group in Table~\ref{tab:omniscenes},\ref{tab:rico360}, and \ref{tab:omniphoto}) and 100 minutes (the second group in Table~\ref{tab:omniscenes},\ref{tab:rico360}, and \ref{tab:omniphoto}) of training time. We compare the 3DGS method based on perspective projection (referred to as 3DGS(P) in the table), the NeRF method (NeRF(P), TensoRF\cite{chen2022tensorf}), and the panoramic customization method (EgoNeRF\cite{choi2023balanced}, ODGS\cite{lee2024odgs}). Tables \ref{tab:omniscenes}, \ref{tab:rico360}, and \ref{tab:omniphoto} show the performance of our method and the state-of-the-art methods on the three datasets. As illustrated in the tables, our method outperforms other methods in almost all scenes, particularly in the SSIM metric. This demonstrates that the novel view rendered by our method better aligns with human visual perception characteristics, thereby enhancing the immersive experience of the scene. We observe that NeRF and TensoRF, which rely on Cartesian coordinate grids, perform poorly in representing large outdoor scenes and significantly lag behind other methods. EgoNeRF, which employs a spherical grid with balanced sampling, achieves better PSNR and SSIM values but still lags behind state-of-the-art methods in the LPIPS metric. Additionally, NeRF-based methods require longer rendering times, posing significant challenges to meeting real-time requirements.
3DGS with perspective images suffers from severe overfitting due to the lack of overlap between the six faces of the cubemap decomposition\cite{lee2024odgs}. This leads to six distinct optimizations for cube faces in various directions, resulting in quick overfitting. The panoramic customization method ODGS shows suboptimal performance except for our method. The tangent approximation reintroduces errors in 3DGS, which generate a large number of redundant 3D Gaussians in highly distorted areas. This significantly increases GPU memory demand (48G of ODGS v.s. 24 G of ours) and causes unstable training, thereby affecting performance in certain scenes. For instance, ODGS fails on the room 1 scene of the OmniScene dataset (Table \ref{tab:omniscenes}), and no results are reported. Despite this, our method still achieves the best performance. In terms of rendering time, ODGS claims their method reaches real-time performance \cite{lee2024odgs}. However, our method can theoretically be rendered in parallel to reduce training time while achieving comparable rendering speed.\\
\noindent \textbf {Qualitative comparison.} %We show qualitative results of our method and comparative methods in Figure \ref{fig:quali}. It can be observed that the perspective 3DGS loses a lot of details between adjacent faces of the cube. In addition, due to the lack of consistency constraints and optimization, obvious stitching seams are generated at the stitching. In contrast, the ODGS method can ensure the consistency of the whole image due to the full-image optimization, but it will produce blur and artifacts due to unstable training and redundant 3D Gaussians. Our method pays attention to details and achieves consistent optimization, so it performs well on both egocentric and roaming datasets.
We show the qualitative results of our method and comparative methods in Figure \ref{fig:quali}. As illustrated in the figure, the perspective 3DGS method exhibits lower image quality. Additionally, due to the absence of consistency constraints and optimization, obvious stitching seams are generated at the stitching boundaries. In contrast, the ODGS method ensures whole-image consistency through full-image optimization but introduces blur and artifacts due to unstable training and redundant 3D Gaussians.
Our method focuses on preserving fine details while achieving consistent optimization. This dual focus results in superior performance on both egocentric and roaming datasets. Specifically, our approach effectively balances detail preservation and global consistency by transition plane, intra-to-inter face optimization strategy, and cube padding approach, thereby mitigating issues such as stitching seams and artifacts observed in other methods.
\begin{table}[t]
\centering
\caption{Ablation study. The best result for each metric is written in bold.}
\label{tab:abla}
\scalebox{0.7}{
\begin{tabular}{c c c c c c c c c c c}
\hline
\multirow{2}{*}{scene} &\multirow{2}{*}{3DGS(P)}& \multirow{2}{*}{TP} & \multirow{2}{*}{OP} & \multirow{2}{*}{CP}&\multicolumn{3}{c}{10 min}& \multicolumn{3}{c}{100 min}\\ \cline{6-11}
\multicolumn{5}{c}{}&PSNR &SSIM &LPIPS &PSNR &SSIM &LPIPS\\ \hline

\multirow{3}{*}{center}  &\checkmark&  &   & &27.24&0.8364& 0.2887&20.04& 0.6974&0.4237 \\ \cline{2-11} 
&\checkmark  & \checkmark &  &  &28.41 &0.8803 &0.1333&28.87&0.8856&0.1418 \\ \cline{2-11} 
&\checkmark  & \checkmark & \checkmark &  &29.08&\textbf{0.8965}&\textbf{0.0940}&29.62&\textbf{0.9044}&\textbf{0.0797} \\ \cline{2-11}
& \checkmark &\checkmark  & \checkmark & \checkmark &\textbf{29.10}& 0.8957& 0.1041&\textbf{29.72}& 0.9034& 0.0827 \\ \hline

\multirow{3}{*}{\small BeihaiPark}  &\checkmark  &  &  & & 23.39& 0.8126& 0.2600&20.81& 0.7581&0.2898\\ \cline{2-11} 
&\checkmark  & \checkmark &  &  &25.68 &0.8707&0.1549&26.22&0.8808&0.1467\\ \cline{2-11} 
&\checkmark  & \checkmark & \checkmark &  &\textbf{26.18} &\textbf{0.8881}&\textbf{0.1087}&26.60 &0.8945&\textbf{0.0898} \\ \cline{2-11}
& \checkmark &\checkmark  & \checkmark & \checkmark &26.05& 0.8848& 0.1193& \textbf{26.77}& \textbf{0.8953}& 0.0904 \\ \hline

 \multirow{3}{*}{room 3}  &\checkmark  &  &  & & 25.13&0.8860& 0.1554&20.13& 0.8066& 0.3109\\ \cline{2-11} 
&\checkmark  & \checkmark &  &  &24.27 &0.8623&0.1793&25.63&0.8912&0.1492 \\ \cline{2-11} 
&\checkmark  & \checkmark & \checkmark &  &26.11 &0.9011&\textbf{0.1222}&26.31&0.8970&0.1115 \\ \cline{2-11}
& \checkmark &\checkmark  & \checkmark & \checkmark &\textbf{26.22}& \textbf{0.9016}& 0.1256& \textbf{26.56}& \textbf{0.8993}& \textbf{0.1106} \\ \hline
\end{tabular}}
\end{table}
\subsection{Ablation study}
We conduct extensive ablation experiments by randomly selecting one scene from each of the three datasets. Our baseline method is 3DGS(P), which is derived from ODGS\cite{lee2024odgs}. In our ablation study, "TP" denotes the use of the Transition Plane method; "OP" refers to the Optimization Strategy; and "CP" indicates the application of spherical padding to the cube. As shown in Table \ref{tab:abla}, we verify the effectiveness of each component by incrementally adding them in a prioritized manner.

\noindent \textbf{Effectiveness of the Transition Plane.}
%To evaluate the impact of the transition plane module, we remove it from the pipeline and only use the six canonical cube faces for optimization. As shown in Table~\ref{tab:abla}, this results in notable artifacts near the cube face boundaries due to ambiguous gradients across discontinuous viewpoints. The absence of intermediate transition views significantly hampers the consistency of 3D Gaussian updates near seams, leading to blurry or over-smoothed reconstructions in these regions. By contrast, our transition plane introduces additional views from in-between perspectives, effectively regularizing optimization near the seams and improving the structural continuity across cube faces. Quantitative improvements in both PSNR and perceptual metrics confirm the benefit of this component.
As shown in Table~\ref{tab:abla}, the baseline exhibits lower performance due to ambiguous gradients across discontinuous viewpoints. By contrast, our transition plane introduces additional views from in-between perspectives, effectively regularizing optimization near the seams and improving the structural continuity across cube faces. Quantitative improvements in both PSNR and perceptual metrics confirm the benefit of this component.

\noindent \textbf{Effectiveness of the optimization strategy.}
We further add an intra-to-inter face optimization strategy to the baseline single-stage training setting and conduct ablation studies. The baseline directly optimizes all Gaussian parameters using only the rendered cube views without incorporating additional stitching or cross-view alignment mechanisms. As shown in Table \ref{tab:abla}, our proposed strategy achieves significant improvements across all three metrics, particularly in the LPIPS metric. This improvement can be attributed to the two-stage intra-to-inter face optimization process, which first refines parameters on individual cube faces and transition views, and then further optimizes them on the stitched panorama. This approach ensures more globally consistent updates and enhances cross-face consistency, as demonstrated by the reduced perceptual artifacts in the final reconstructions.

\begin{figure}[t]
  \centering
  \includegraphics[width=0.5\textwidth]{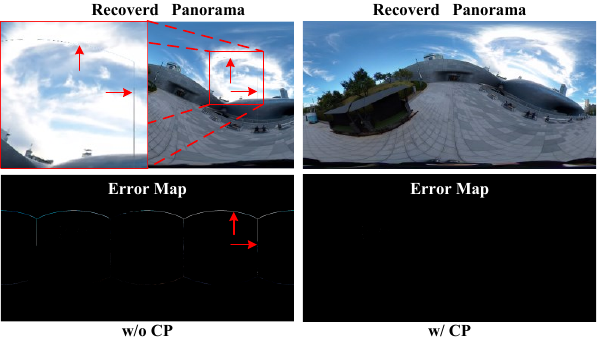} %1.png是图片文件的相对路径 
  \caption{Effectiveness of cube padding (CP).} %caption是图片的标题
  \label{fig:error}
\end{figure}

\noindent \textbf{Effectiveness of the Cube Padding.}
As observed in Figure~\ref{fig:error}, sampling discontinuities near the cube face borders result in visible seams during ERP transformation. The padding, implemented via spherical sampling extensions, ensures smoother coordinate transitions. As shown in the table, although it may slightly harm SSIM and LPIPS, it yields more seamless panoramic images and further stabilizes 3D Gaussian updates in boundary regions. 
\section{Conclusion}
In this paper, we propose a novel framework that bridges the continuous omnidirectional 3D scene reconstruction with perspective 3D Gaussian Splatting. To bridge the domain gap, we introduce a transition plane method that relieves the ambiguity when splatting to the cube faces. Moreover, we propose a two-stage optimization strategy to address the overfitting and degeneration problem of perspective 3D Gaussian Splatting. Additionally, we introduce an efficient and camera-free spherical sampling technique to erase the visible stitching seams. Extensive experiments on indoor and outdoor, egocentric, and roaming datasets demonstrate that our method outperforms current state-of-the-art approaches. 

{\small
\bibliographystyle{IEEEtran}
\bibliography{main}
}

\vfill

% \appendix

\end{document}